\pdfoutput=1
\documentclass[9pt,twocolumn,twoside]{opticajnl}
\journal{opticajournal}

\setboolean{shortarticle}{false}

\usepackage{float}
\usepackage{stfloats}
\graphicspath{{figures/}}

\title{Physics-Informed Untrained Learning for RGB-Guided Superresolution Single-Pixel Hyperspectral Imaging}
\dates{}
\doi{}

\author[1]{Hao Zhang}
\author[1]{Bilige Xu}
\author[1]{Lichen Wei}
\author[2,3]{Xu Ma}
\author[1,*]{Wenyi Ren}

\affil[1]{College of Science, Northwest Agriculture \& Forestry University, Yangling 712100, China}
\affil[2]{Key Laboratory of Photoelectronic Imaging Technology and System, School of Optics and Photonics, Beijing Institute of Technology, Beijing, China}
\affil[3]{State Key Laboratory of Digital Manufacturing Equipment and Technology, Huazhong University of Science and Technology, Wuhan 430074, China}
\affil[*]{renovelhuman@gmail.com}

\begin{abstract}
Single-pixel imaging (SPI) offers a cost-effective route to hyperspectral acquisition but struggles to recover high-fidelity spatial and spectral details under extremely low sampling rates—a severely ill-posed inverse problem. While deep learning has shown potential, existing data-driven methods demand large-scale pretraining datasets that are often impractical in hyperspectral imaging. To overcome this limitation, we propose an end-to-end physics-informed framework that leverages untrained neural networks and RGB guidance for joint hyperspectral reconstruction and super-resolution without any external training data. The framework comprises three physically grounded stages: (1) a Regularized Least-Squares method with RGB-derived Grayscale Priors (LS-RGP) that initializes the solution by exploiting cross-modal structural correlations; (2) an Untrained Hyperspectral Recovery Network (UHRNet) that refines the reconstruction through measurement consistency and hybrid regularization; and (3) a Transformer-based Untrained Super-Resolution Network (USRNet) that upsamples the spatial resolution via cross-modal attention, transferring high-frequency details from the RGB guide. Extensive experiments on benchmark datasets demonstrate that our approach significantly surpasses state-of-the-art algorithms in both reconstruction accuracy and spectral fidelity. Moreover, a proof-of-concept experiment using a physical single-pixel imaging system validates the framework's practical applicability, successfully reconstructing a 144-band hyperspectral data cube at a mere 6.25\% sampling rate. The proposed method thus provides a robust, data-efficient solution for computational hyperspectral imaging.
\end{abstract}

\setboolean{displaycopyright}{false}

\begin{document}
\maketitle

\section{Introduction}

Single-pixel imaging (SPI) is an emerging computational imaging technique that has garnered considerable interest due to its advantages in non-visible spectral regions and low-light environments \cite{tian_joint_2023,deng_high-efficiency_2023,zhang_vgennet_2023}. By employing a single-pixel detector with structured illumination patterns, SPI enables cost-effective hyperspectral imaging where array detectors are prohibitively expensive or unavailable \cite{yang_underwater_2021,ribes_towards_2020}.

However, reconstructing high-quality hyperspectral images under extremely low sampling rates remains a formidable challenge. Traditional algorithms such as compressed sensing (CS) often suffer from severe performance degradation below 5\% sampling, leading to artifacts, noise, and spectral distortion \cite{he_ghost_2018,zhang_computational_2021}. Although deep learning methods have recently improved reconstruction quality, they typically require massive annotated datasets for supervised training—a requirement that is difficult to meet in hyperspectral imaging due to the scarcity of aligned RGB-hyperspectral pairs \cite{bian_residual-based_2020}. Moreover, models trained on specific datasets often generalize poorly to new scenes or hardware configurations. To address these issues, we present a novel framework that synergistically integrates physical measurement models, RGB guidance, and untrained neural networks to achieve high-fidelity reconstruction without external training data.

\begin{figure*}[!t]
  \centering
  \includegraphics[width=0.98\textwidth]{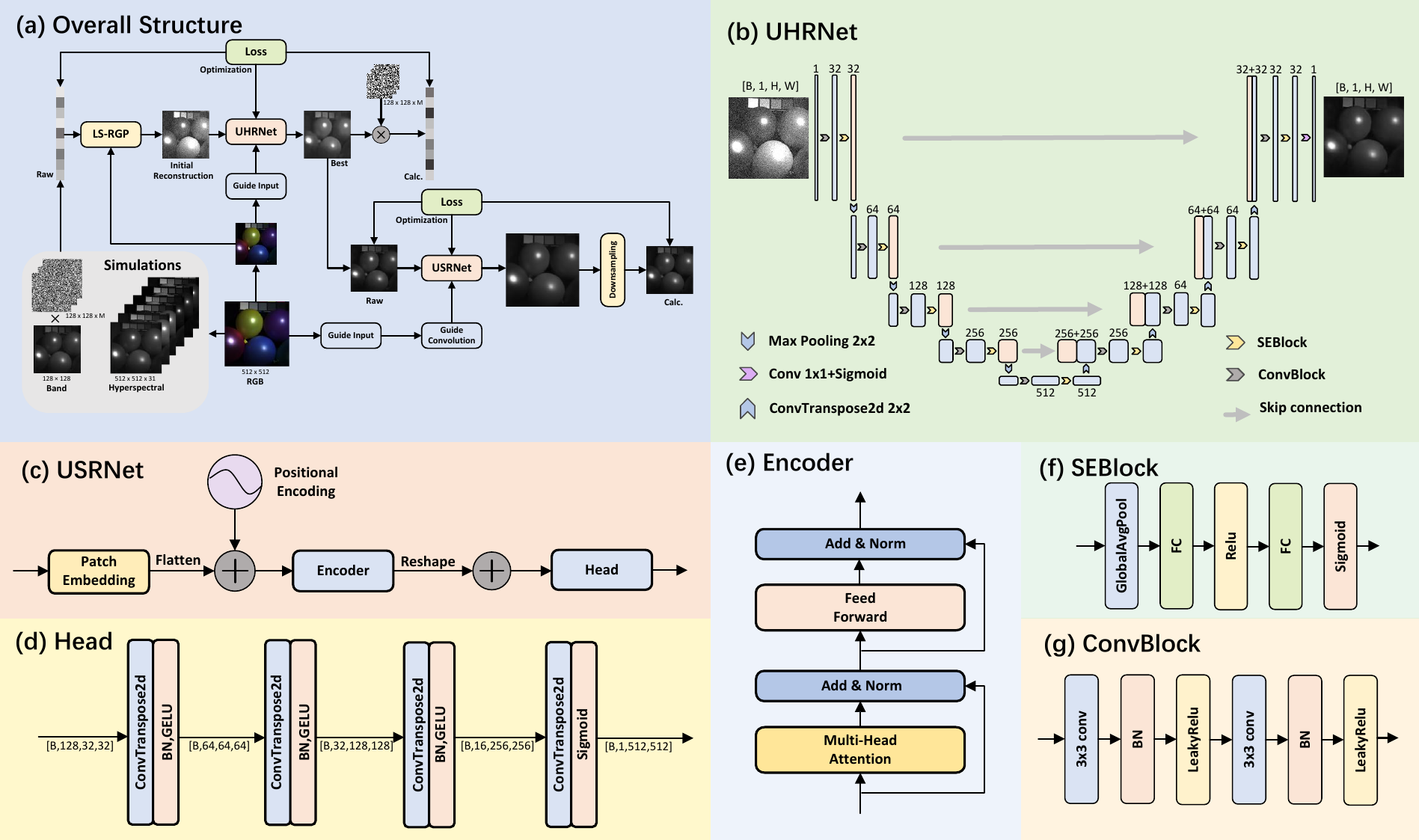}
  \caption{The overall architecture of the proposed RGB-guided hyperspectral reconstruction framework. (a) End-to-end pipeline integrating SPI physics, RGB guidance, and untrained neural networks. (b) UHRNet: RGB-guided hyperspectral recovery network. (c) USRNet: transformer-based hyperspectral super-resolution network. (d) Head module for feature mapping. (e) Encoder with multi-head attention. (f) SEBlock for channel-wise attention. (g) ConvBlock with convolution and normalization layers.}
  \label{fig:fig1_total_network}
\end{figure*}

\subsection{Related Work}

Our work builds upon advances in three key areas: single-pixel hyperspectral imaging(SPHI), RGB-guided reconstruction, and untrained neural networks.

\subsubsection{Single-Pixel Hyperspectral Imaging}
SPI has evolved from early correlation-based methods such as differential ghost imaging (DGI) \cite{ferri_differential_2010, gong_method_2010} to advanced compressed sensing and deep learning approaches. CS algorithms like TVAL3 \cite{li_efficient_2010} exploit image sparsity to reduce sampling requirements but often incur high computational costs and struggle to recover fine textures at sampling rates below 5\% \cite{noauthor_computational_2018,zhai_computational_2019}. Recent deep learning methods learn the mapping from 1D measurements to 2D images \cite{wang_single-pixel_2022,chang_self-supervised_2023}, yet their supervised nature limits flexibility in data-scarce scenarios.

\subsubsection{RGB-Guided Hyperspectral Reconstruction}
RGB images provide high-resolution spatial structures that can compensate for the low spatial resolution of hyperspectral sensors \cite{cai_mst_2022}. Existing approaches often treat this as a spectral super-resolution or fusion problem \cite{wang_single-pixel_2023}, but they typically assume a fixed spectral response and require extensive paired training data. In SPI, effectively fusing 1D compressed measurements with 2D RGB priors remains challenging without relying on large-scale pretraining.

\subsubsection{Untrained Neural Networks}
Untrained neural networks (UNNs), also known as Deep Image Prior (DIP), optimize network weights on a single test instance using the network structure as a prior \cite{ulyanov_deep_2020,wang_phase_2020,liu_computational_2021}. This paradigm is advantageous when ground truth data is unavailable. Recent works have applied UNNs to standard SPI \cite{peng_high-fidelity_2023,li_urnet_2023}, but extending them to SPHI is challenging due to high dimensionality and spectral consistency requirements. Our work extends the UNN paradigm by incorporating explicit RGB guidance and physical measurement constraints.

\subsection{Main Contributions}

To address the dual challenges of extremely low sampling rates and data scarcity, we propose an end-to-end framework with the following key contributions:

\begin{itemize}[wide=0\parindent]
    \item A unified physics-informed framework that synergistically integrates SPI measurement models, RGB guidance, and untrained neural networks, enabling high-quality hyperspectral reconstruction at extremely low sampling rates without large-scale pretraining data.
    \item A Regularized Least Squares initialization with RGB Prior (LS-RGP) that analytically incorporates RGB-derived structural information to provide a robust starting point for optimization.
    \item A dual-stage untrained network architecture comprising an RGB-guided Hyperspectral Recovery Network (UHRNet) and a Transformer-based Super-resolution Network (USRNet), which enforce physical consistency and employ cross-modal attention to progressively refine spatial and spectral details.
    \item Extensive experiments on benchmark datasets demonstrating that our method significantly outperforms state-of-the-art optimization-based and deep learning-based methods (including MST++ and PYFINETUNE) in both spatial reconstruction quality and spectral accuracy.
    \item Validation through a real-world SPHI experiment, successfully reconstructing a 144-band data cube from physical measurements at a 6.25\% sampling rate, confirming practical feasibility.
\end{itemize}

\section{Methods}

\subsection{Overall Framework}
As illustrated in Fig. \ref{fig:fig1_total_network}, our framework consists of three core stages. First, the SPI measurement and preliminary reconstruction stage (LS-RGP) acquires one-dimensional measurement signals under extremely low sampling rates using random binary patterns and performs an initial reconstruction by integrating RGB-derived grayscale priors. Second, the RGB-guided hyperspectral recovery stage employs UHRNet to refine the preliminary reconstruction through iterative optimization that enforces measurement consistency, Fourier regularization, perceptual loss, sharpness loss, and spatial smoothness loss. Third, the hyperspectral super-resolution stage utilizes USRNet, a transformer-based network, to upscale the spatial resolution by transferring high-frequency details from the RGB guide via cross-modal attention. The entire process operates without external training data, fully leveraging physical constraints and RGB guidance to achieve high-quality reconstruction under severe undersampling.

\subsection{SPI Forward Model and Initialization}
In SPI, a sequence of $ M $ patterns illuminates the scene, and the reflected intensity is recorded by a single-pixel detector. For a hyperspectral image $ X \in \mathbb{R}^{H \times W \times B} $, each pattern $ P_m \in \mathbb{R}^{H \times W} $ yields a scalar measurement:
\begin{equation}
y_m = \langle X, P_m \rangle + \epsilon, \quad m=1,2,\dots,M,
\end{equation}
where $ \epsilon $ denotes measurement noise. Vectorizing the $ b $-th band as $ x_b \in \mathbb{R}^{H \times W} $, the measurement for that band is:
\begin{equation}
y_b = P x_b + \epsilon.
\end{equation}
Under extremely low sampling rates ($ M \ll H \times W $), the system is severely underdetermined. To incorporate prior knowledge, we convert the RGB guidance image $ I_{\text{RGB}} $ to a grayscale prior $ I_{\text{gray}} $ via luminance extraction:
\begin{equation}
I_{\text{gray}} = 0.299 \cdot I_{\text{R}} + 0.587 \cdot I_{\text{G}} + 0.114 \cdot I_{\text{B}}.
\end{equation}
The preliminary reconstruction is obtained by solving the regularized least-squares problem:
\begin{equation}
\hat{x}_b = \arg\min_{x_b} \|y_b - P x_b\|_2^2 + \lambda\|x_b - I_{\text{gray}}\|_2^2,
\end{equation}
which has the closed-form solution:
\begin{equation}
\hat{x}_b = (P^\top P + \lambda I)^{-1}(P^\top y_b + \lambda I_{\text{gray}}).
\end{equation}
This LS-RGP initialization provides a stable starting point but still contains noise and missing details, which are addressed in the subsequent stages.

In practice, the measurement patterns are binary random masks generated by thresholding a normally distributed matrix, facilitating hardware implementation using Digital Micromirror Devices (DMDs).

\subsection{RGB-Guided Hyperspectral Recovery Network (UHRNet)}

\subsubsection{Network Architecture}
UHRNet is built upon a U-Net backbone with skip connections and incorporates a Squeeze-and-Excitation (SE) block for channel attention. The encoder progressively downsamples the input feature maps via convolution and max-pooling, capturing multi-scale representations. The decoder upsamples through transposed convolutions, with skip connections preserving fine spatial details. The SE block recalibrates channel-wise features by first computing global average pooling:
\begin{equation}
z_c = \frac{1}{H \times W} \sum_{i=1}^H \sum_{j=1}^W X_{ij,c},
\end{equation}
then applying two fully-connected layers with a Sigmoid activation to generate attention weights $ s_c $, which rescale the input features:
\begin{equation}
X_{\text{refined}} = X \cdot \sigma(s_c).
\end{equation}
This design enhances the network's ability to focus on informative spectral-spatial features.

\subsubsection{Loss Functions}
The training of UHRNet minimizes a composite loss function that enforces physical consistency and perceptual quality:
\begin{equation}
\begin{aligned}
\mathcal{L} =& \mathcal{L}_{\text{meas}} + \lambda_{\text{phy}} (\mathcal{L}_{\text{Fourier}} + \mathcal{L}_{\text{sharp}} + \mathcal{L}_{\text{smooth}}) + \lambda_{\text{rgb}} \mathcal{L}_{\text{percep}},
\end{aligned}
\end{equation}
where\\
\begin{equation}
\begin{aligned} \mathcal{L}_{\text{meas}} & = \| \mathcal{M}(X_{\text{rec}}) - y \|_2^2, \\
 \mathcal{L}_{\text{Fourier}} & = \|\mathcal{F}(X_{\text{rec}})\|_1, \\
\mathcal{L}_{\text{sharp}} & = \sum_{i,j} |\mathcal{L}_{\text{lap}} X_{\text{rec}}(i,j)|, \\
\mathcal{L}_{\text{smooth}} & = \sum_{i,j} \left( |\nabla_h X_{\text{rec}}(i,j)| + |\nabla_v X_{\text{rec}}(i,j)| \right), \\
\mathcal{L}_{\text{percep}} & = \frac{1}{C_l H_l W_l} \|\Phi_l(X_{\text{rec}}) - \Phi_l(I_{\text{RGB}})\|_2^2.
\end{aligned}
\end{equation}
Here, $ \mathcal{M}(\cdot) $ simulates the SPI measurement process, $ \mathcal{F}(\cdot) $ denotes the 2D Fourier transform, $ \mathcal{L}_{\text{lap}} $ is the Laplacian operator, and $ \Phi_l(\cdot) $ represents features from the $ l $-th layer of a pretrained VGG network. The weights $ \lambda_{\text{phy}} $ and $ \lambda_{\text{rgb}} $ balance the contributions.

\begin{figure*}[!tbp]
  \centering
  \includegraphics[width=0.98\textwidth]{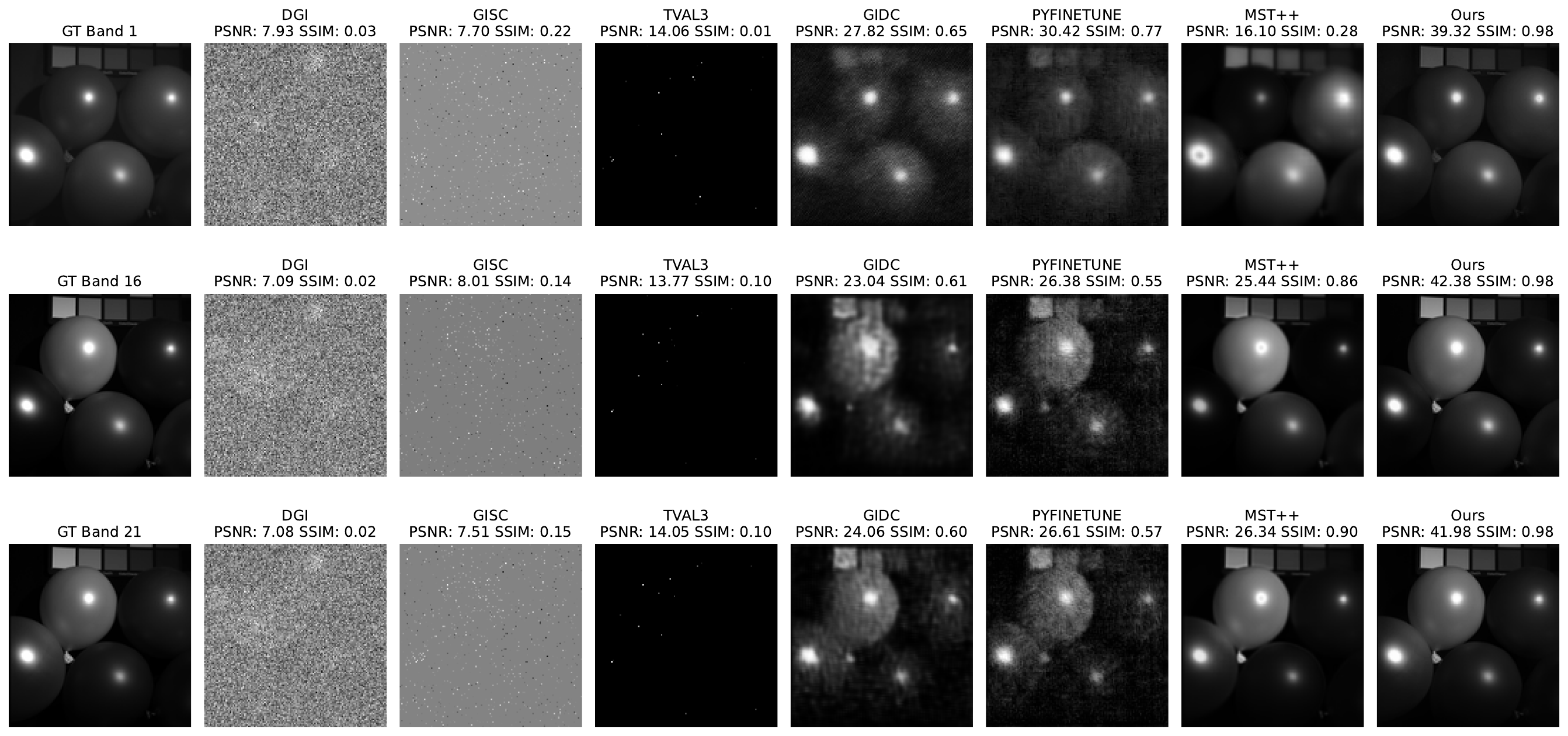}
  \caption{Comparison of hyperspectral reconstruction quality across different methods for Bands 1, 16, and 21. Grayscale heatmaps visualize spatial fidelity. From left to right: Ground Truth (GT), DGI, GISC, TVAL3, GIDC, PYFINETUNE, MST++, and Ours. PSNR and SSIM values are displayed. Our method consistently achieves higher metrics and superior visual quality.}
  \label{fig:fig2_multiband_comparison}
\end{figure*}

\subsection{Hyperspectral Super-Resolution Network (USRNet)}

\subsubsection{Network Architecture}
USRNet is a transformer-based network designed to upsample the low-resolution hyperspectral image ($128\times128$) to high resolution ($512\times512$) using RGB guidance. The input is first embedded via convolutional layers. Positional encoding is added to retain spatial information. The core transformer module employs multi-head self-attention to capture long-range dependencies across spatial and spectral dimensions. For input features $ X \in \mathbb{R}^{B \times N \times C} $, query, key, and value matrices are computed as:
\begin{equation}
Q = XW^Q,\quad K = XW^K,\quad V = XW^V,
\end{equation}
and multi-head attention outputs are concatenated and linearly transformed:
\begin{equation}
\text{MultiHead}(X) = \mathrm{Concat}(\text{head}_1, \dots, \text{head}_h) W^O.
\end{equation}
Residual connections and layer normalization stabilize training. An RGB guidance path extracts luminance information from the high-resolution RGB image, downscales it to match the hyperspectral feature scale, and fuses it with the transformer features via channel-wise concatenation. Finally, transposed convolutions progressively upsample to the target resolution.

\subsubsection{Loss Functions}
The super-resolution loss combines multiple terms to ensure spatial and spectral fidelity:
\begin{equation}
\begin{aligned}
\mathcal{L}_{\text{SR}} =& \mathcal{L}_{\text{down}} + \lambda_{\text{phy}} (\mathcal{L}_{\text{Fourier}} + \mathcal{L}_{\text{TV}}) + \lambda_{\text{rgb}} (\mathcal{L}_{\text{percep}} + \mathcal{L}_{\text{SSIM}}),
\end{aligned}
\end{equation}
where
\begin{equation}
\begin{aligned} \mathcal{L}_{\text{down}} & = \| \text{downsample}(X_{\text{HR}}) - X_{\text{LR}} \|_2^2, \\
 \mathcal{L}_{\text{TV}} & = \sum_{i,j} \left( |\nabla_h X_{\text{HR}}(i,j)| + |\nabla_v X_{\text{HR}}(i,j)| \right), \\
 \mathcal{L}_{\text{SSIM}} & = 1 - \text{SSIM}(X_{\text{HR}}, I_{\text{RGB}}).
\end{aligned}
\end{equation}
Here, $ \mathcal{L}_{\text{down}} $ enforces consistency with the low-resolution input, $ \mathcal{L}_{\text{TV}} $ promotes edge-preserving smoothness, and SSIM loss aligns structural details with the RGB guide.

\section{Simulation Experiments and Analysis}

We evaluate our method on the CAVE dataset \cite{yasuma_generalized_2010}, containing 32 hyperspectral scenes with 31 bands (400–700 nm). Images are normalized to [0,1]. Comparisons are made against state-of-the-art algorithms: MST++ \cite{cai_mst_2022}, PYFINETUNE \cite{wang_single-pixel_2022}, DGI \cite{ferri_differential_2010}, GIDC \cite{wang_far-field_2022}, GISC \cite{gong_high-resolution_nodate}, TVAL3 \cite{li_efficient_2010}, and DIP \cite{ulyanov_deep_2020}. Implementation uses PyTorch on an NVIDIA RTX 4090D GPU, with the AdamW optimizer (initial learning rate $4\times10^{-3}$, weight decay $1\times10^{-4}$), and learning rate decay by 0.8 every 3000 iterations for 10,000 iterations total. No external pretraining is used. Metrics: Peak Signal-to-Noise Ratio (PSNR), Structural Similarity Index (SSIM), and Spectral Angle Mapper (SAM).

\subsection{Reconstruction Quality Across Bands}

Figure~\ref{fig:fig2_multiband_comparison} compares reconstruction results for Bands 1, 16, and 21. Visually and quantitatively, our method outperforms others, e.g., in Band 16 achieving PSNR 42.38 dB and SSIM 0.98 versus MST++ (25.44 dB, 0.86). The recovered images exhibit sharper edges, better texture preservation, and closer resemblance to the ground truth.

\begin{figure*}[!tbp]
  \centering
  \includegraphics[width=0.98\textwidth]{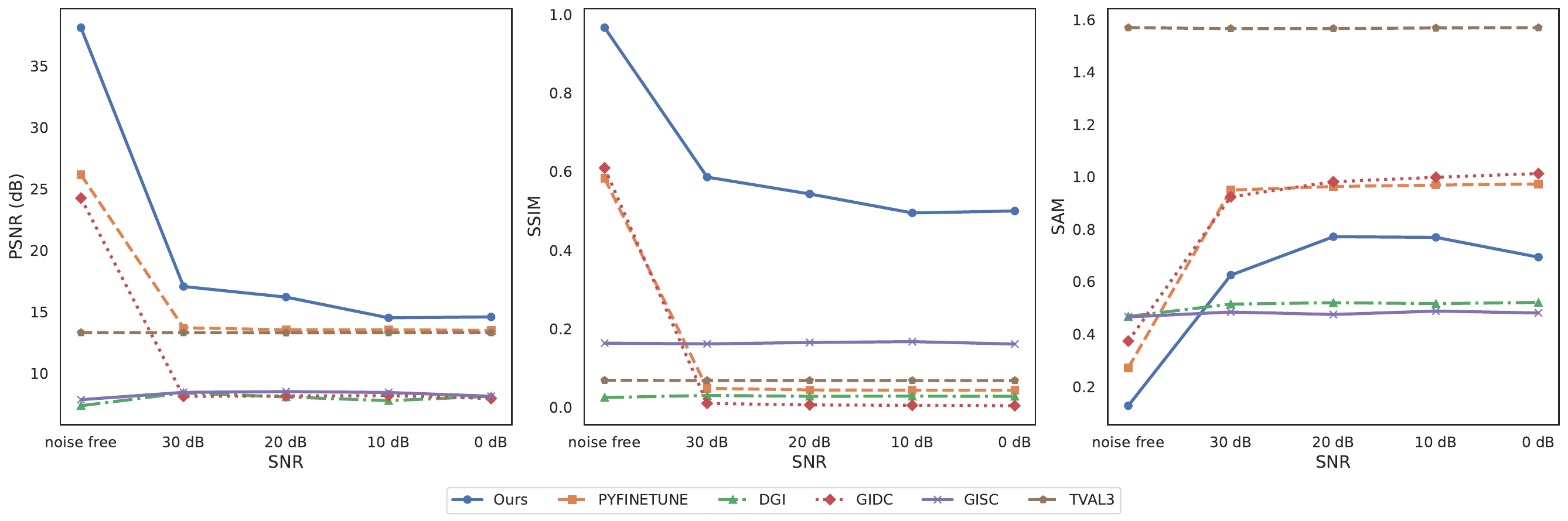}
  \caption{Performance under different SNR conditions at 6.25\% sampling rate. (a) Average PSNR, (b) average SSIM, and (c) average SAM for each method. Our approach shows significantly better noise resilience.}
  \label{fig:fig4_noise_psnr_ssim_sam}
\end{figure*}

\subsection{Overall Quantitative Performance}

\begin{figure}[!htbp]
  \centering
  \includegraphics[width=0.95\columnwidth]{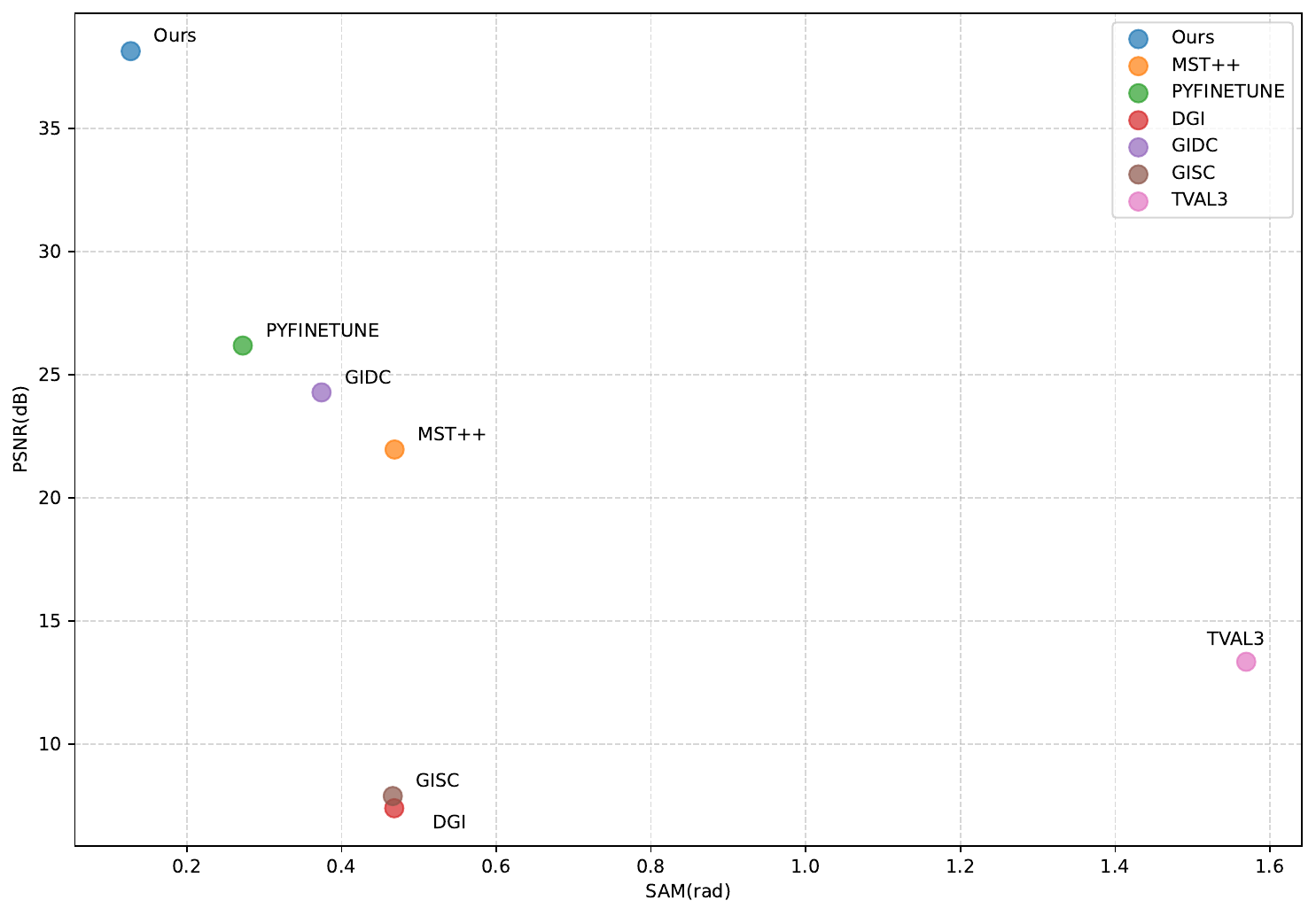}
  \caption{Overall performance comparison in terms of PSNR and SAM. Our method achieves the best trade-off, indicating superior reconstruction fidelity and spectral preservation.}
  \label{fig:fig3_sam_vs_psnr}
\end{figure}

Figure~\ref{fig:fig3_sam_vs_psnr} presents a scatter plot of PSNR versus SAM across all test bands. Our method achieves the highest average PSNR (38.14 dB) and lowest SAM (0.13 rad), substantially outperforming the second-best, PYFINETUNE (26.18 dB, 0.27 rad). This demonstrates excellence in both spatial detail recovery and spectral accuracy, attributed to the effective use of RGB priors that constrain the solution space and mitigate noise overfitting.

\subsection{Robustness to Noise}

Figure~\ref{fig:fig4_noise_psnr_ssim_sam} evaluates performance under varying SNR levels. At 0 dB SNR, other methods degrade severely (PSNR <10 dB, SSIM $\approx$0), while our method maintains PSNR $\sim$15 dB and SSIM >0.5. This robustness stems from the hybrid regularization and RGB guidance, which collectively suppress noise while preserving structural information.

\subsection{Impact of Pattern Count}

\begin{figure}[!htbp]
  \centering
  \includegraphics[width=0.95\columnwidth]{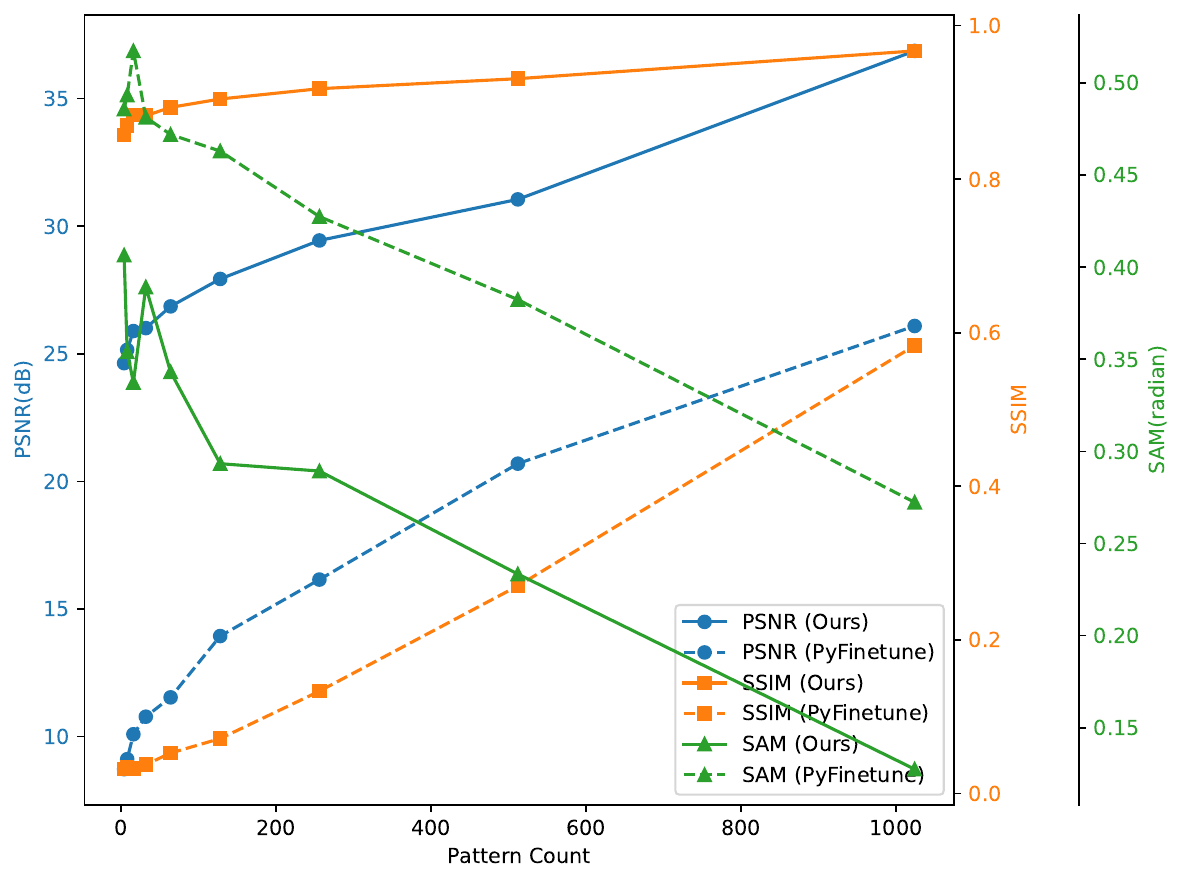}
  \caption{Effect of the number of measurement patterns on performance metrics (PSNR, SSIM, SAM) for our method and PYFINETUNE.}
  \label{fig:fig5_psnr_ssim_sam_all_in_one}
\end{figure}

\begin{figure}[!htbp]
  \centering
  \includegraphics[width=0.8\columnwidth]{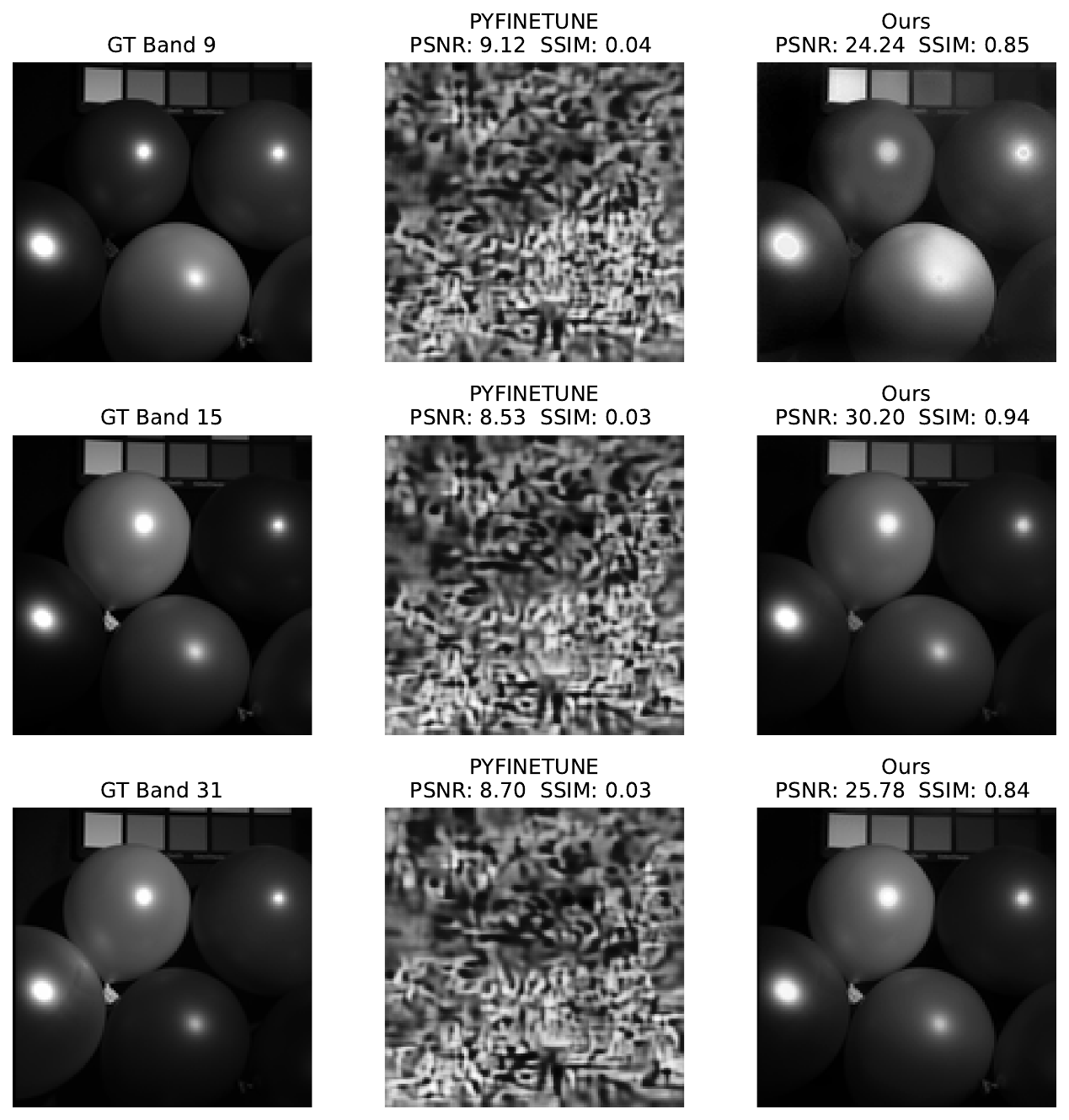}
  \caption{Visual comparison for Bands 9, 15, and 31 using only 4 patterns. Our method yields clearer textures and higher fidelity metrics.}
  \label{fig:fig6_comparison}
\end{figure}

Figure~\ref{fig:fig5_psnr_ssim_sam_all_in_one} shows how performance varies with the number of measurement patterns. Even with only 4 patterns, our method attains PSNR >25 dB and SSIM $\approx$0.8, demonstrating strong adaptability to extreme undersampling. As patterns increase to 1024, PSNR and SSIM improve further while SAM decreases below 0.2 rad, indicating stable and accurate recovery across sampling conditions. Figure~\ref{fig:fig6_comparison} provides a visual comparison for Bands 9, 15, and 31 using merely 4 patterns. Our method retains more spatial details and suppresses noise better than PYFINETUNE, corroborated by higher PSNR and SSIM values.

\subsection{Spectral Accuracy}

\begin{figure}[!htbp]
  \centering
  \includegraphics[width=0.95\columnwidth]{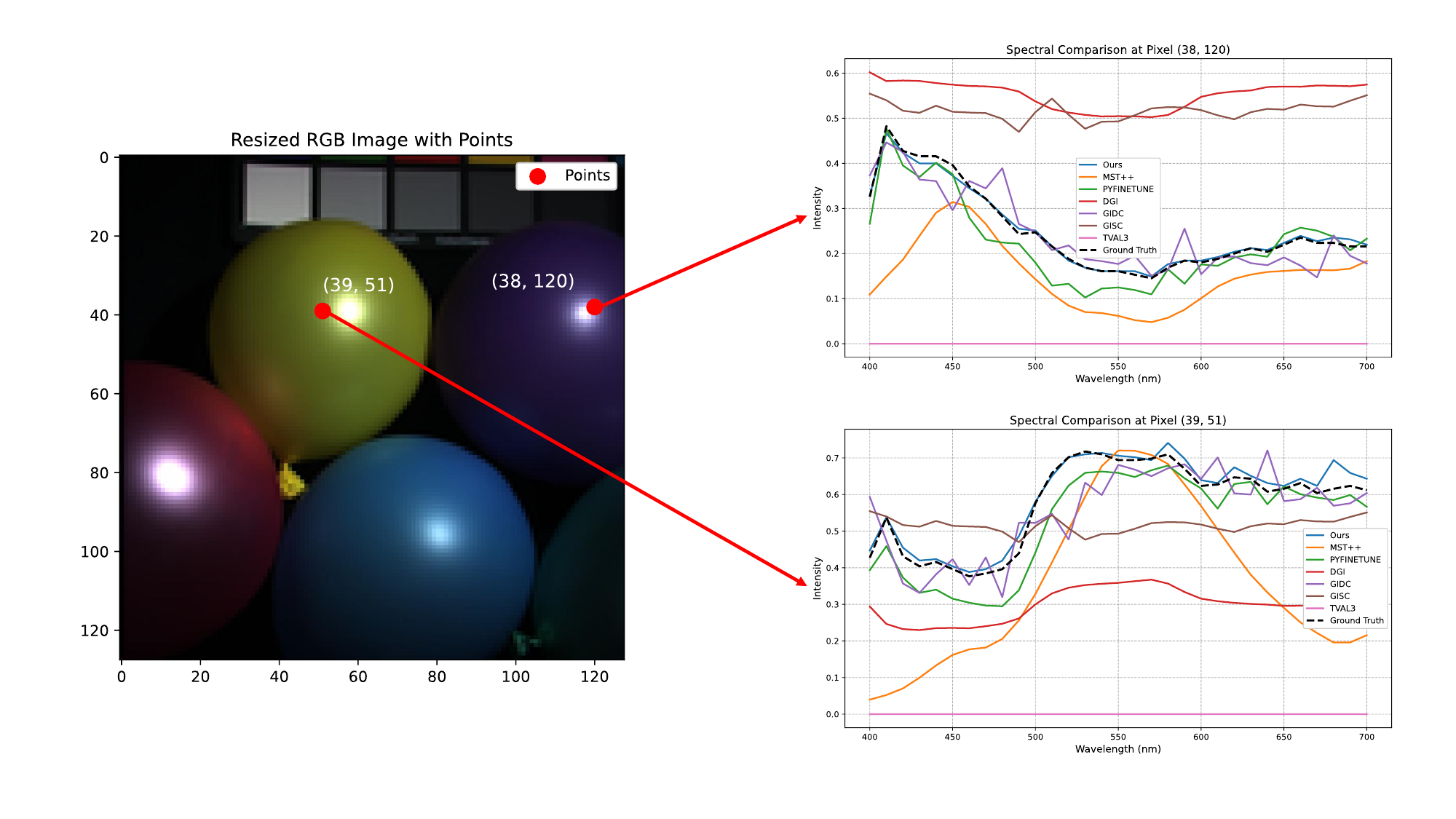}
  \caption{Spectral recovery comparison for two pixels (marked in the RGB image) at 6.25\% sampling. Our method (red) aligns best with the ground truth (blue), exhibiting the lowest SAM values.}
  \label{fig:fig7_spectral_comparison}
\end{figure}

Figure~\ref{fig:fig7_spectral_comparison} plots spectral curves for two representative pixels. Our reconstructed spectra closely match the ground truth, with SAM values as low as 0.0350 rad and 0.0351 rad, significantly lower than other methods. This confirms precise spectral feature recovery, essential for applications like material identification.

\begin{figure*}[!tbp]
  \centering
  \includegraphics[width=0.98\textwidth]{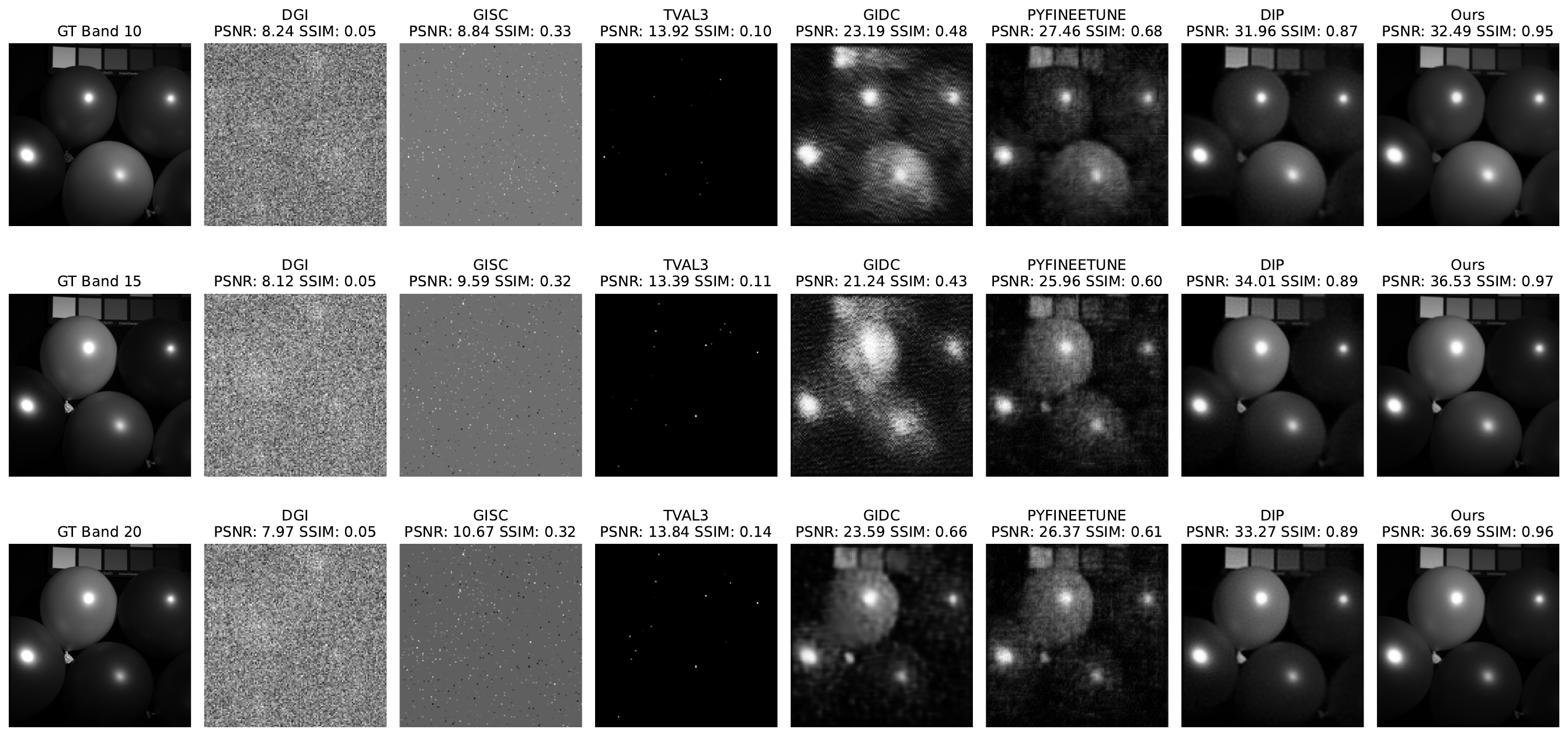}
  \caption{Super-resolution reconstruction for Bands 10, 15, and 20. Our method delivers superior visual quality and objective metrics.}
  \label{fig:fig8_multiband_comparison_sr}
\end{figure*}

\subsection{Super-Resolution Performance}

Table \ref{tab:sr_results} summarizes super-resolution results. Our method achieves the highest PSNR (35.01 dB), SSIM (0.9571), and lowest SAM (0.1326 rad), outperforming all competitors. This underscores its ability to enhance spatial resolution while preserving spectral integrity.

\begin{table}[!htbp]
\centering
\caption{Super-resolution performance comparison. Our method excels across all metrics.}
\label{tab:sr_results}
\resizebox{\columnwidth}{!}{%
\begin{tabular}{lcccc}
\hline
\textbf{Method} & \textbf{PSNR (dB)} & \textbf{SSIM} & \textbf{SAM (°)} & \textbf{SAM (rad)} \\
\hline
Ours            & \textbf{35.0147} & \textbf{0.9571} & \textbf{7.5949} & \textbf{0.1326} \\
DGI             & 8.3223  & 0.0519 & 27.0787 & 0.4726 \\
GIDC            & 22.7902 & 0.5758 & 22.9502 & 0.4006 \\
GISC            & 9.9916  & 0.3472 & 27.2265 & 0.4752 \\
TVAL3           & 13.2984 & 0.0991 & 89.9110 & 1.5692 \\
PYFINETUNE      & 26.2854 & 0.6361 & 14.5766 & 0.2544 \\
DIP             & 31.4653 & 0.8713 & 13.9991 & 0.2443 \\
\hline
\end{tabular}%
}
\end{table}

Visual results for Bands 10, 15, and 20 are shown in Fig. \ref{fig:fig8_multiband_comparison_sr}, where our method produces sharper details and more accurate colors.

\subsection{Local Detail Analysis}

\begin{figure}[!htbp]
  \centering
  \includegraphics[width=0.95\columnwidth]{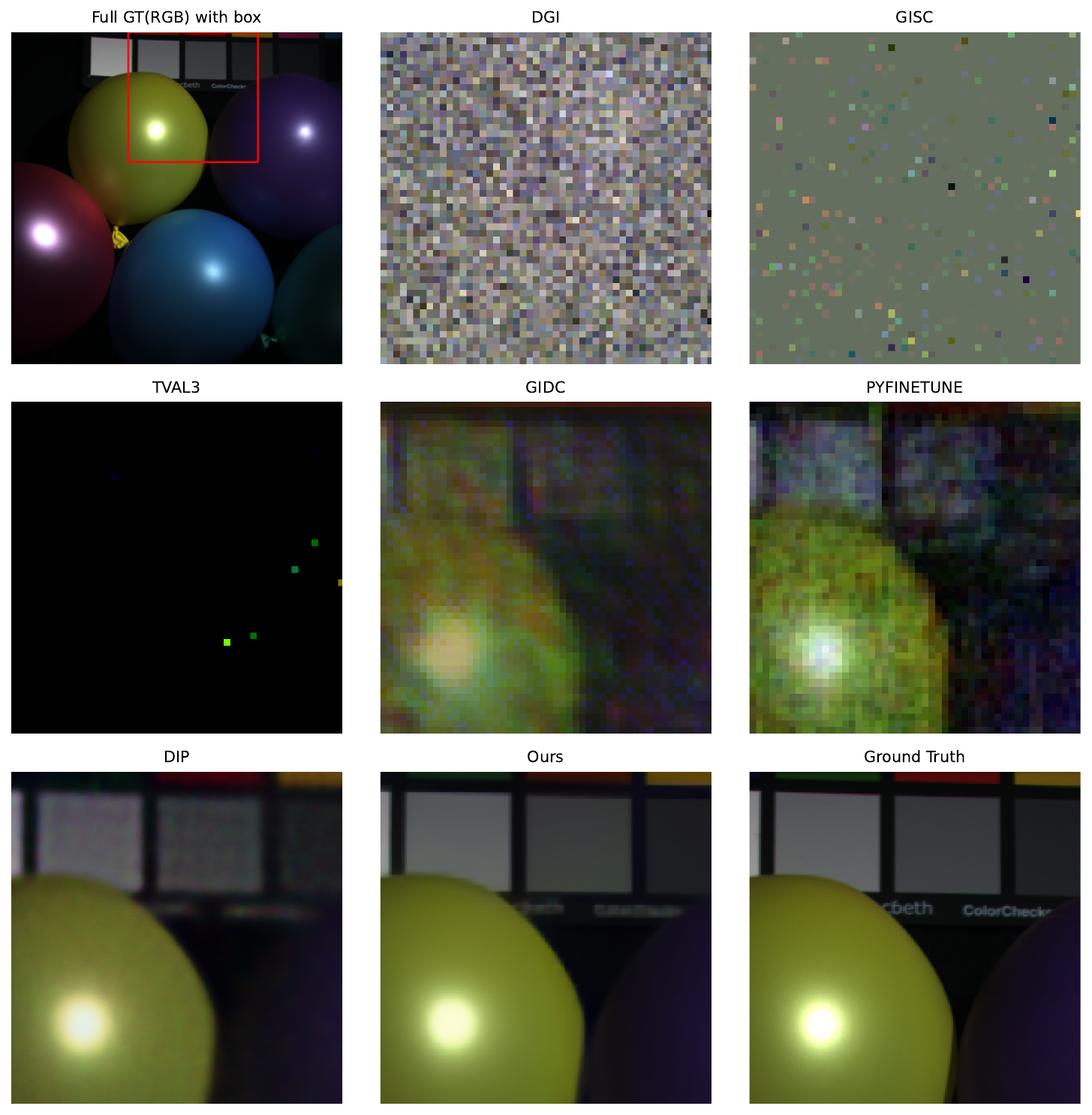}
  \caption{ROI comparison: the RGB image with marked ROI (top left) and enlarged reconstructions from each method versus ground truth. Our result exhibits the finest detail preservation.}
  \label{fig:fig9_roi_comparison}
\end{figure}

Figure~\ref{fig:fig9_roi_comparison} zooms into a region of interest (ROI). Our reconstruction reveals clearer textures and more faithful spectral content compared to other methods, with minimal deviation from the ground truth.

\subsection{Ablation Studies}

\subsubsection{Hyperspectral Recovery Components}

Table \ref{tab:ablation_psnr_ssim_sam} ablates loss terms in UHRNet. Removing any component (e.g., Fourier regularization, perceptual loss) leads to noticeable drops in PSNR, SSIM, and SAM, confirming that each contributes to the overall performance. The perceptual loss, in particular, is crucial for maintaining visual realism.

\begin{table}[!htbp] 
\centering
\caption{Ablation study for RGB-guided hyperspectral recovery. Each loss term positively impacts performance.}
\label{tab:ablation_psnr_ssim_sam}
\resizebox{\columnwidth}{!}{%
\begin{tabular}{lcccc}
\hline
\textbf{Configuration} & \textbf{PSNR (dB)} & \textbf{SSIM} & \textbf{SAM (°)} & \textbf{SAM (rad)} \\
\hline
Full model            & \textbf{38.1406} & \textbf{0.9670} & \textbf{7.3064} & \textbf{0.1275} \\
w/o measurement loss  & 25.5383 & 0.8616 & 21.2551 & 0.3710 \\
w/o Fourier reg.      & 36.5002 & 0.9546 & 12.5095 & 0.2183 \\
w/o sharpness loss    & 29.2406 & 0.8379 & 29.4664 & 0.5143 \\
w/o smoothness loss   & 37.3611 & 0.9602 & 12.3805 & 0.2161 \\
w/o perceptual loss   & 14.6182 & 0.5841 & 19.8794 & 0.3470 \\
\hline
\end{tabular}%
}
\end{table}

\subsubsection{Super-Resolution Components}

Table \ref{tab:sr_ablation} examines USRNet. Replacing transformer blocks with standard layers (“no\_transblocks”) causes a large performance decline (PSNR drops to 32.43 dB), highlighting the importance of long-range dependency modeling. Removing RGB guidance (“no\_rgb\_guidance”) or downsampling consistency loss (“no\_downsample”) also degrades results, validating the design of cross-modal fusion and multi-scale constraints.

\begin{table}[!htbp]
\centering
\caption{Ablation study for super-resolution network. The transformer architecture and RGB guidance are essential for high-quality upsampling.}
\label{tab:sr_ablation}
\resizebox{\columnwidth}{!}{%
\begin{tabular}{lcccc}
\hline
\textbf{Configuration} & \textbf{PSNR (dB)} & \textbf{SSIM} & \textbf{SAM (°)} & \textbf{SAM (rad)} \\
\hline
Full model (Baseline)   & \textbf{35.01} & \textbf{0.9571} & \textbf{7.59} & \textbf{0.1326} \\
\hline
\textit{Architecture:} \\
 no transformer blocks  & 32.43 & 0.9400 & 10.87 & 0.1896 \\
 shallow (depth=4)      & 34.64 & 0.9558 & 7.64 & 0.1334 \\
 fewer heads (heads=2)  & 33.64 & 0.9457 & 8.66 & 0.1512 \\
 narrow (dim=64)        & 34.77 & 0.9560 & 7.58 & 0.1323 \\
 no positional encoding & 34.68 & 0.9561 & 7.64 & 0.1334 \\
 no RGB guidance        & 34.31 & 0.9545 & 7.75 & 0.1352 \\
\hline
\textit{Loss functions:} \\
 no downsampling loss   & 21.09 & 0.8201 & 31.83 & 0.5555 \\
 no SSIM loss           & 33.20 & 0.9374 & 7.68 & 0.1340 \\
 no TV loss             & 33.80 & 0.9310 & 12.58 & 0.2196 \\
 no Fourier loss        & 34.90 & 0.9566 & 7.69 & 0.1342 \\
 no perceptual loss     & 32.45 & 0.9400 & 10.78 & 0.1882 \\
\hline
\end{tabular}%
}
\end{table}

\section{Real-World Experimental Validation}

To demonstrate practical applicability, we implemented a physical SPHI system and tested the proposed framework on real measurements.

\subsection{Experimental Setup}
\begin{figure}[!htbp]
  \centering
  \includegraphics[width=\columnwidth]{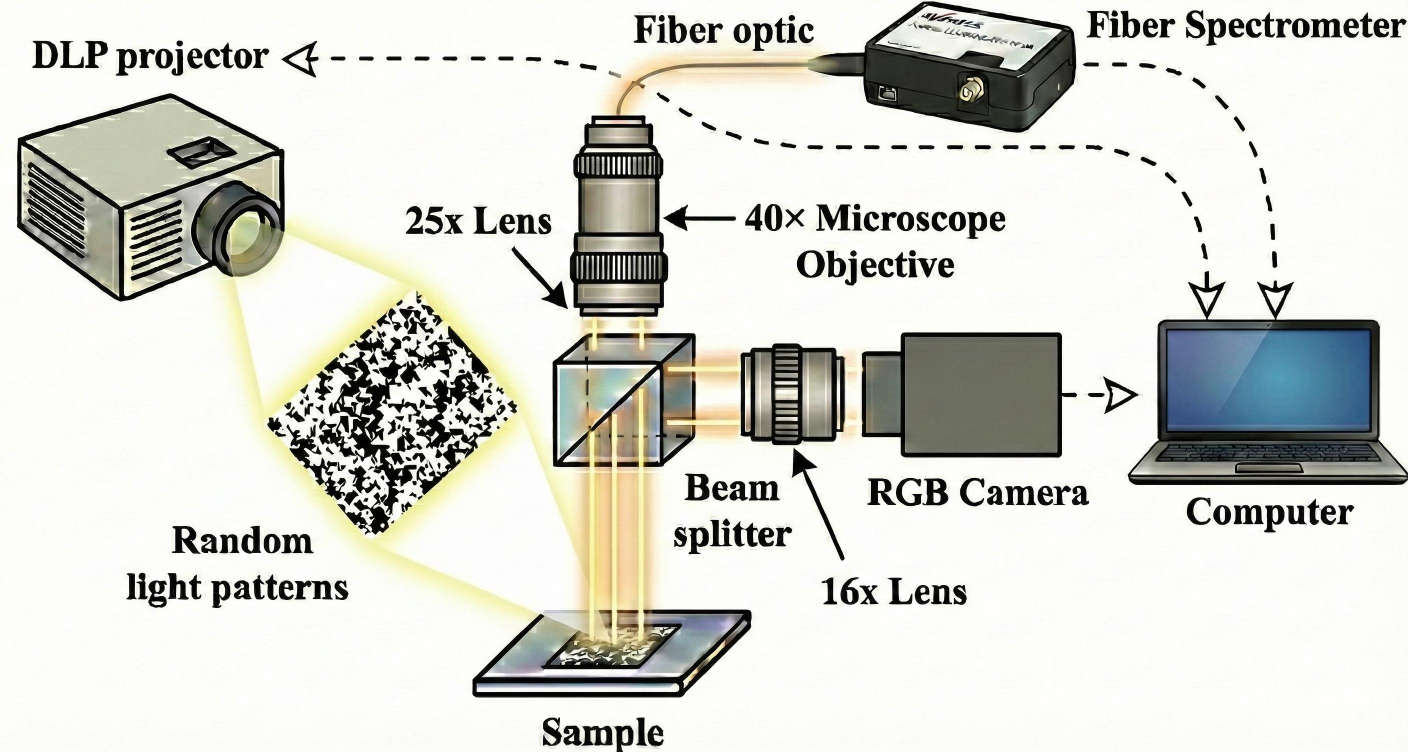}
  \caption{Schematic of the real-world SPHI system. Broadband illumination from the DLP projector is structured by random patterns. Reflected light from the target enters a beam splitter; the left path is captured by the RGB camera (equipped with a 16$\times$ lens), and the straight path is directed to the spectrometer via a 25$\times$ lens and a 40$\times$ microscope objective.}
  \label{fig:real_schematic}
\end{figure}

\begin{figure}[!htbp]
  \centering
  \includegraphics[width=0.95\columnwidth, height=0.7\textheight, keepaspectratio]{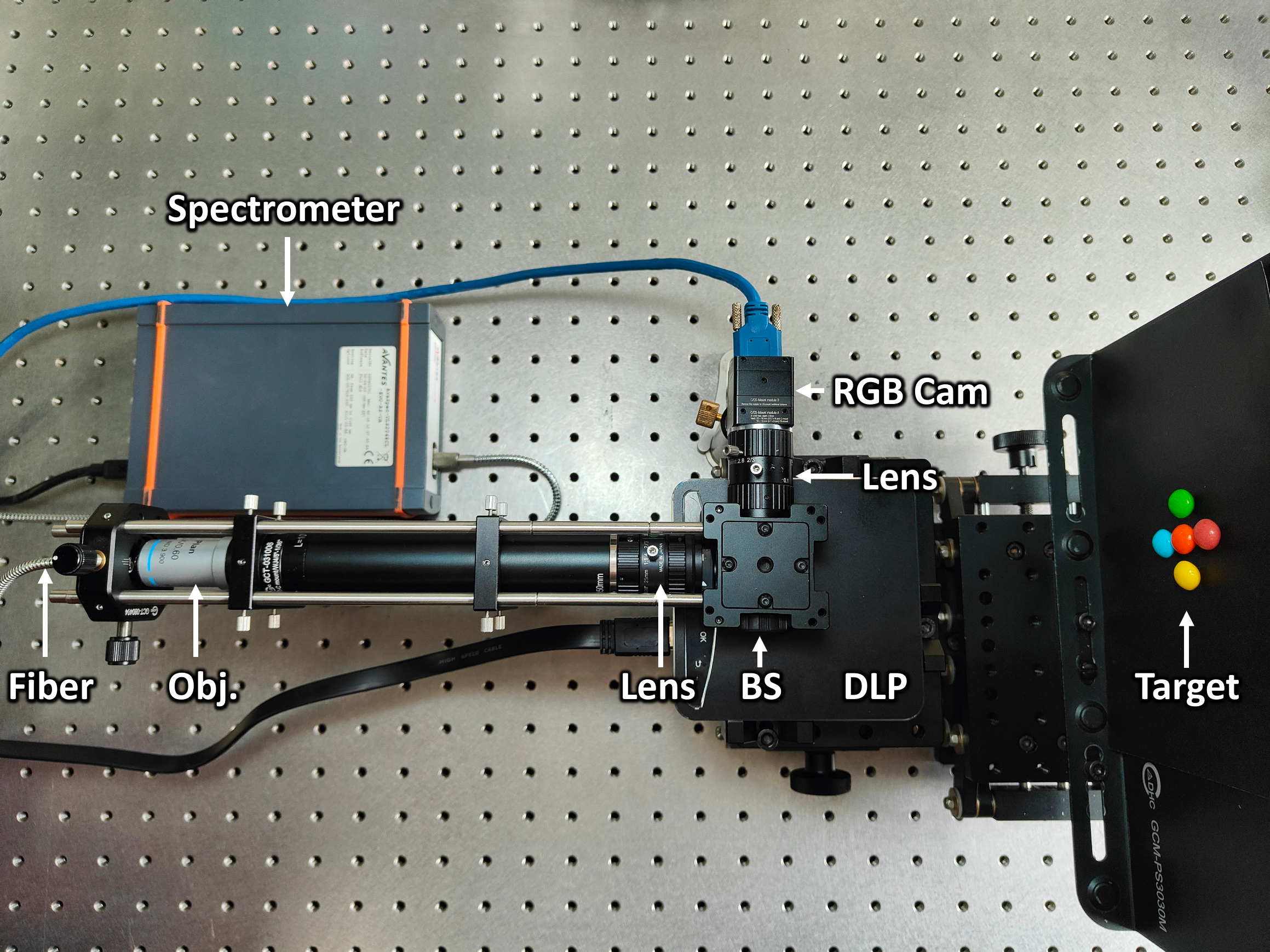}
  \caption{Photograph of the experimental setup showing key components: DLP projector, beam splitter (BS), RGB camera (RGB Cam), collection lens (Lens), microscope objective (Obj.), and fiber spectrometer (Spec.).}
  \label{fig:real_photo}
\end{figure}

The system (Figs. \ref{fig:real_schematic} and \ref{fig:real_photo}) includes a DLP projector for structured illumination, a fiber spectrometer for spectral detection, an RGB camera for spatial guidance, and a computer for data acquisition. In this optical path, light reflects from the target to a beam splitter. The left path leads to the RGB camera, which is equipped with a 16$\times$ lens in front of it. The straight path directs light to the spectrometer, passing sequentially through a 25$\times$ lens and a 40$\times$ microscope objective before reaching the fiber probe. Binary random patterns of size $128\times128$ are projected; $M=1024$ patterns yield a sampling rate of 6.25\%. The spectrometer records 144 bands from 380 nm to 720 nm (2.4 nm interval). The RGB camera captures a high-resolution color image for guidance. Key parameters are listed in Table \ref{tab:real_params}.

\begin{table}[!htbp]
\centering
\caption{Key parameters of the real-world experimental system.}
\label{tab:real_params}
\begin{tabular}{ll}
\hline
\textbf{Parameter} & \textbf{Value} \\
\hline
Spatial resolution & $128 \times 128$ pixels \\
Number of patterns ($M$) & 1024 \\
Sampling rate ($\beta$) & 6.25\% \\
Spectral range & 380--720 nm \\
Number of bands ($B$) & 144 \\
Spectral interval & $\approx$ 2.4 nm \\
Pattern type & Random binary \\
Projector & DLP, 100 W \\
Spectrometer & Avantes ULS2048CL-EVO \\
RGB camera & XIMEA MQ042CG, $2048 \times 2048$ \\
\hline
\end{tabular}
\end{table}

\subsection{Reconstruction Results}
Following the same three-stage pipeline, we reconstruct a $128 \times 128 \times 144$ hyperspectral data cube from the physical measurements. Despite low sampling and real-world noise, the reconstructed images exhibit clear spatial structures across bands.

\subsubsection{Spectral Fidelity}
\begin{figure}[!htbp]
  \centering
  \includegraphics[width=\columnwidth]{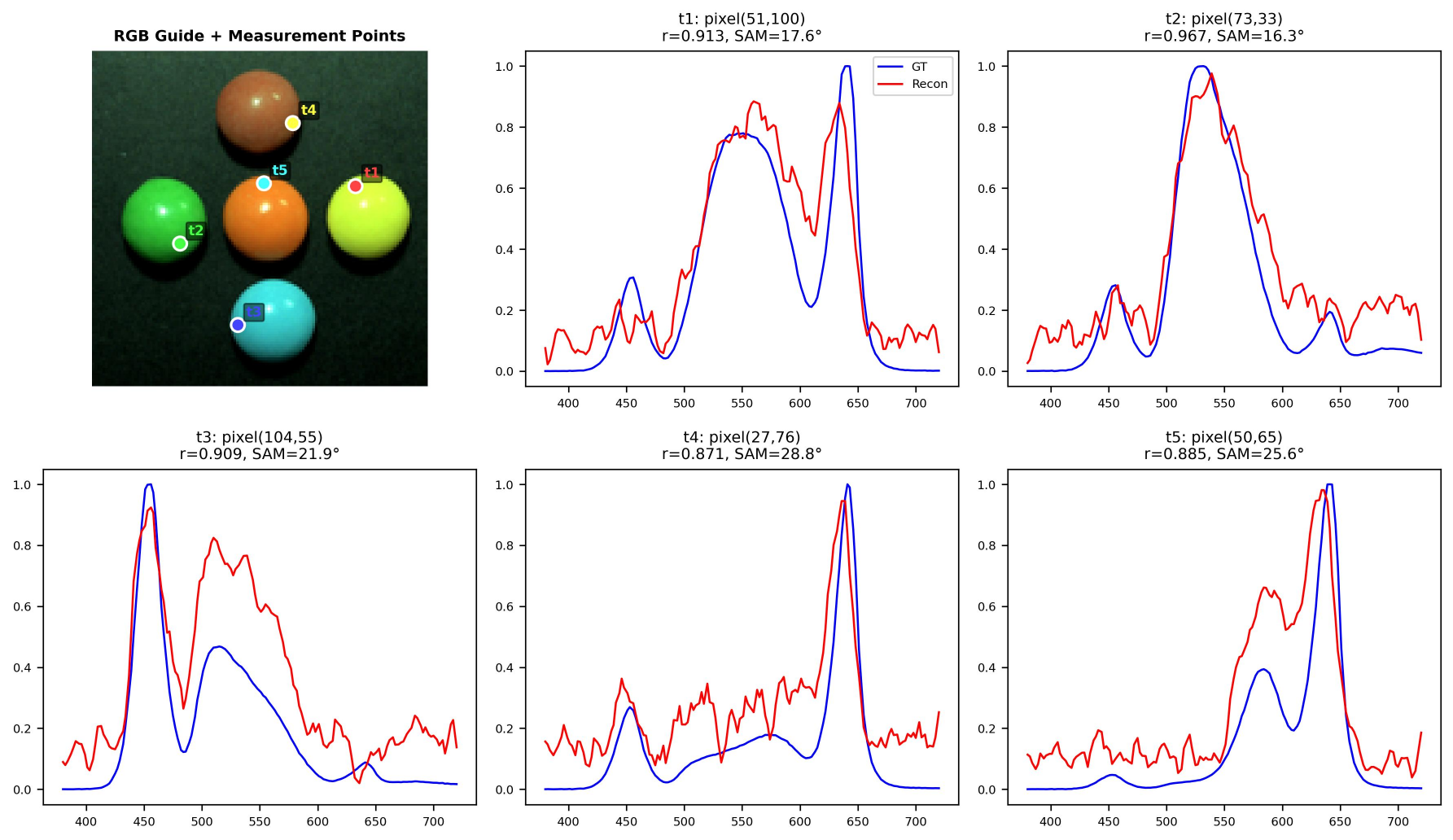}
  \caption{Spectral fidelity analysis. (a) RGB image with five target locations marked. (b)–(f) Ground truth (blue) vs. reconstructed spectra (red) for each target, with Pearson correlation $r$ and SAM annotated.}
  \label{fig:real_spectra}
\end{figure}

The target scene comprises five colored Skittles candies. Point measurements from a fiber spectrometer provide ground truth spectra for each candy. We identify the best-matching pixel in the reconstruction via spectral angle minimization and compare spectra. Figure~\ref{fig:real_spectra} shows the results; after Savitzky–Golay (SG) smoothing, the mean Pearson correlation across targets is $r = 0.909$ and mean SAM is $22.0^\circ$ (Table~\ref{tab:real_metrics}). The SG filtering improves correlation by +2.61\% and reduces SAM by $1.5^\circ$, confirming that spectral features are accurately recovered.

\begin{table}[!htbp]
\centering
\caption{Quantitative spectral fidelity metrics for the five targets. SG smoothing enhances correlation and reduces SAM.}
\label{tab:real_metrics}
\begin{tabular}{lcccc}
\hline
\textbf{Target} & \textbf{Pixel} & $r$ (raw) & $r$ (SG) & \textbf{SAM (SG)} \\
\hline
Target 1 & (51, 100) & 0.899 & 0.913 & 17.6$^\circ$ \\
Target 2 & (73, 33)  & 0.946 & 0.967 & 16.3$^\circ$ \\
Target 3 & (104, 55) & 0.887 & 0.909 & 21.9$^\circ$ \\
Target 4 & (27, 76)  & 0.813 & 0.871 & 28.8$^\circ$ \\
Target 5 & (50, 65)  & 0.869 & 0.885 & 25.6$^\circ$ \\
\hline
\textbf{Mean} & — & 0.883 & \textbf{0.909} & \textbf{22.0}$^\circ$ \\
\hline
\end{tabular}
\end{table}

\subsubsection{Hyperspectral Data Cube Visualization}
\begin{figure}[!ht]
  \vspace{-10pt} 
  \centering
  \includegraphics[width=\columnwidth]{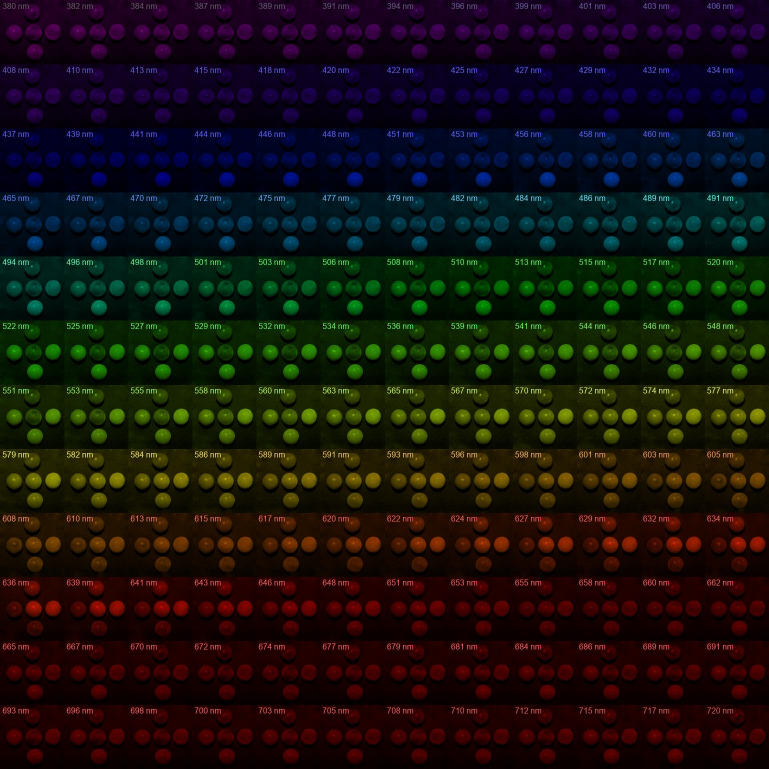}
  \caption{Visualization of the reconstructed $128 \times 128 \times 144$ hyperspectral data cube from real measurements, covering 380--720 nm.}
  \label{fig:real_datacube}
\end{figure}

The full data cube is visualized in Fig.~\ref{fig:real_datacube}. False-color composites and single-band images confirm spatially and spectrally consistent recovery, demonstrating the framework's practicality.

\section{Conclusion}

We have presented a physics-informed untrained learning framework for RGB-guided super-resolution SPHI. By integrating LS-RGP initialization, UHRNet, and USRNet, the method achieves high-fidelity reconstruction under extremely low sampling rates without requiring external training data. Extensive simulations show superior performance over state-of-the-art algorithms in spatial and spectral metrics. A real-world experiment validates practical feasibility, successfully reconstructing a 144-band hyperspectral cube at 6.25\% sampling. The framework effectively addresses the data-scarcity challenge in computational hyperspectral imaging.

Limitations include longer reconstruction time due to iterative optimization and sensitivity to misalignment between RGB and SPI modalities. Future work will focus on acceleration via meta-learning and robust cross-modal registration to handle parallax and alignment errors.


\begin{backmatter}

\bmsection{Funding}
This work was supported by: Shaanxi Fundamental Science Research Project for Mathematics and Physics (23JSY-013); Key Research and Development Program of Shaanxi (2024NC-YBXM-215); Shaanxi Natural Science Fundamental Research Program (2024JC-YBQN-0051); Major Science and Technology Special Program of Yunnan Province (202402AE090005); Shaanxi Qinchuangyuan Industrial Innovation Cluster "Four-Chain" Integration Project (2025CY-JJQ-21); Key Industrial Innovation Chain Project of Shaanxi Province (2024NC-ZDCYL-05-01).

\bmsection{Disclosures}
The authors declare no conflicts of interest.

\bmsection{Data availability}
Data underlying the results presented in this paper are not publicly available at this time but may be obtained from the authors upon reasonable request.

\end{backmatter}

\bibliography{references}

\bibliographyfullrefs{references}

\end{document}